\documentclass{article}

    \usepackage[final]{neurips_2020}

\PassOptionsToPackage{numbers, compress}{natbib}
\usepackage[dvipsnames]{xcolor}

\usepackage{graphicx}
\usepackage{subfigure}
\usepackage{neurips_2020}

\usepackage[utf8]{inputenc} 
\usepackage[T1]{fontenc}    
\usepackage{hyperref}       
\usepackage{url}            
\usepackage{booktabs}       
\usepackage{nicefrac}       
\usepackage{microtype}      
\usepackage{wrapfig}

\usepackage{amssymb}
\usepackage{amsthm}
\usepackage{amsmath,amsfonts,bm}
\usepackage{accents}
\usepackage[customcolors]{hf-tikz}
\usepackage{scrextend}
\usepackage{comment}
\usepackage{gensymb}
\usepackage{float} 
\usepackage{overpic}

\newcommand{\lcom}[1]{\textcolor{blue}{[luke: #1]}}

\theoremstyle{definition}
\newtheorem{definition}{Definition}
\newtheorem{theorem}{Theorem}

\title{On Linear Identifiability of Learned Representations}

\author{%
 Geoffrey Roeder\thanks{Corresponding authors. Research done while the first author was an intern at Google Brain.} \\
  Princeton University\\
  Google Brain \\
  \texttt{roeder@princeton.edu} \\
   \And
   Luke Metz \\
   Google Brain \\
   \texttt{lmetz@google.com} \\
   \AND
   Diederik P. Kingma$^*$\\
   Google Brain \\
   \texttt{durk@google.com} \\
}

\def\eqref#1{equation~\ref{#1}}
\def\Eqref#1{Equation~\ref{#1}}

\def\1{\bm{1}}

\def\rvc{{\mathbf{c}}}

\def\rve{{\mathbf{e}}}
\def\rvf{{\mathbf{f}}}
\def\rvg{{\mathbf{g}}}
\def\rvh{{\mathbf{h}}}

\def\rvw{{\mathbf{w}}}
\def\rvx{{\mathbf{x}}}
\def\rvy{{\mathbf{y}}}
\def\rvz{{\mathbf{z}}}

\def\vz{{\bm{z}}}

\DeclareMathAlphabet{\mathsfit}{\encodingdefault}{\sfdefault}{m}{sl}
\SetMathAlphabet{\mathsfit}{bold}{\encodingdefault}{\sfdefault}{bx}{n}

\def\gU{{\mathcal{U}}}

\newcommand{\E}{\mathbb{E}}

\newcommand{\bb}[1]{\mathbf{#1}}

\newcommand{\linsim}{\stackrel{\text{\scalebox{.75}{L}}}{\sim}}

\newcommand{\bS}{\bb{S}}

\newcommand{\bT}{\boldsymbol{\theta}}

\newcommand{\pT}{p_{\bT}}

\newcommand{\pD}{p_{\mathcal{D}}}

\newcommand{\rvfT}{\rvf_{\bT}}
\newcommand{\rvgT}{\rvg_{\bT}}

\newcommand{\ga}{g_{AR}}

\newcommand{\xtk}{\rvx_{t+k}}
\newcommand{\cpaparams}{\boldsymbol{\phi}}

\newcommand{\nlm}{P_{\theta}^h(w)}
\newcommand{\score}{s_{\theta}(w,h)}

\definecolor{skyblue_ggr}{RGB}{86,180,233}
\definecolor{orange_ggr}{RGB}{230,159,0}
\definecolor{reddishpurple}{RGB}{204,121,167}
\definecolor{vermillion}{RGB}{213,94,0}
\definecolor{blusihgreen}{RGB}{0,158,115}

\newcommand{\dcol}{skyblue_ggr!60} 
\newcommand{\ccol}{orange_ggr!60} 
\newcommand{\constcol}{blusihgreen!60}

\newcommand{\dcolbox}[2]
{\tikzmarkin[set fill color=\dcol, set border color=\dcol]{#1}(.05,-0.125)(-.05,0.3)#2\tikzmarkend{#1}}

\newcommand{\ccolbox}[2]
{\tikzmarkin[set fill color=\ccol, set border color=\ccol]{#1}(.05,-0.125)(-.05,0.3)#2\tikzmarkend{#1}}

\newcommand{\constcolbox}[2]
{\tikzmarkin[set fill color=\constcol, set border color=\constcol]{#1}(.05,-0.125)(-.05,0.3)#2\tikzmarkend{#1}}

\newcommand{\largeconstcolbox}[2]
{\tikzmarkin[set fill color=\constcol, set border color=\constcol]{#1}(0.0,-0.50)(0.0,0.450)#2\tikzmarkend{#1}}

\newcommand{\ubar}[1]{\underaccent{\bar}{#1}}

\begin{document}

\maketitle

\begin{abstract}
Identifiability is a desirable property of a statistical model: it implies that the true model parameters may be estimated to any desired precision, given sufficient computational resources and data. 
We study identifiability in the context of representation learning: discovering nonlinear data representations that are optimal with respect to some downstream task. 
When parameterized as deep neural networks, such representation functions typically lack identifiability in parameter space, because they are overparameterized by design.
In this paper, building on recent advances in nonlinear ICA, we aim to rehabilitate identifiability by showing that a large family of discriminative models are in fact identifiable in {\it function} space, up to a linear indeterminacy.
Many models for representation learning in a wide variety of domains have been identifiable in this sense, including text, images and audio, state-of-the-art at time of publication.
We derive sufficient conditions for linear identifiability and provide empirical support for the result on both simulated and real-world data.
\end{abstract}

\section{Introduction}
An increasingly common technique in modern computational statistics is to learn high-dimensional representations of data using deep neural networks to improve performance on a down-stream tasks. 
In this paradigm, training a model reduces to fine-tuning the learned representations for optimal performance on a particular sub-task~\citep{erhan2010does}.
Deep neural networks, as flexible function approximators, have been surprisingly successful in discovering high-dimensional representations for use in downstream tasks in diverse problem domains, including image classification ~\citep{sharif2014cnn}, text generation ~\citep{radford2018improving,devlin2018bert}, speech generation, and sequential decision making ~\citep{oord2018representation}.

When using learned representations for downstream tasks, it is useful when the learned representations are reproducible: that every time we learn the representation function, it is approximately the same, regardless of small deviations in the initialization of the parameters or the optimization procedure. 
A rigorous way to achieve reproducibility is to choose a model whose representation function is \emph{identifiable} in function space. Informally speaking, identifiability in function space is achieved when, in the limit of infinite data, there exists a single, global optimum in function space. 

Recent advances in the theory of nonlinear ICA have proven strong identifiability results \citep{hyvarinen2018nonlinear,khemakhem2019variational,khemakhem2020ice,sorrenson2020disentanglement}.
These latter works have provided a deeper understanding of and sufficient conditions for identifiability of generative models of data. 
We aim to bridge the gap between this theory and the dramatic success of discriminative models for representation learning (e.g., \citep{henaff2019data}). See Section \ref {sec:related_work} for a discussion of connections with these earlier works.

In short, we make the following contributions:
\begin{itemize}
    \item In Section \ref{sec:gdl}, we describe a general discriminative model family, defined by its canonical mathematical form, which generalizes many supervised, self-supervised, and contrastive learning frameworks.
    \item In Section \ref{sec:identifiability}, we prove that learned representations in this family have an asymptotic property desirable for representation learning: equality up to a linear transformation.
    \item In Section \ref{sec:lit}, we show that this family includes a number of highly performant models, state-of-the-art at publication for their problem domains, including CPC \citep{oord2018representation}, BERT \citep{devlin2018bert}, and GPT-2 and GPT-3 \citep{radford2018improving,radford2019language,brown2020language}.
    \item  Section {\ref{sec:experiments}} investigates the realizable regime of {\it finite} data and {\it partial} estimation. We present strong evidence that representations learned by members of the identifiable model family approach equality up to a linear transformation as function of dataset size, neural network capacity, and optimization progress.
\end{itemize}

\section{Model Family}
\begin{figure}[t]
  \begin{center}
    \includegraphics[width=\linewidth]{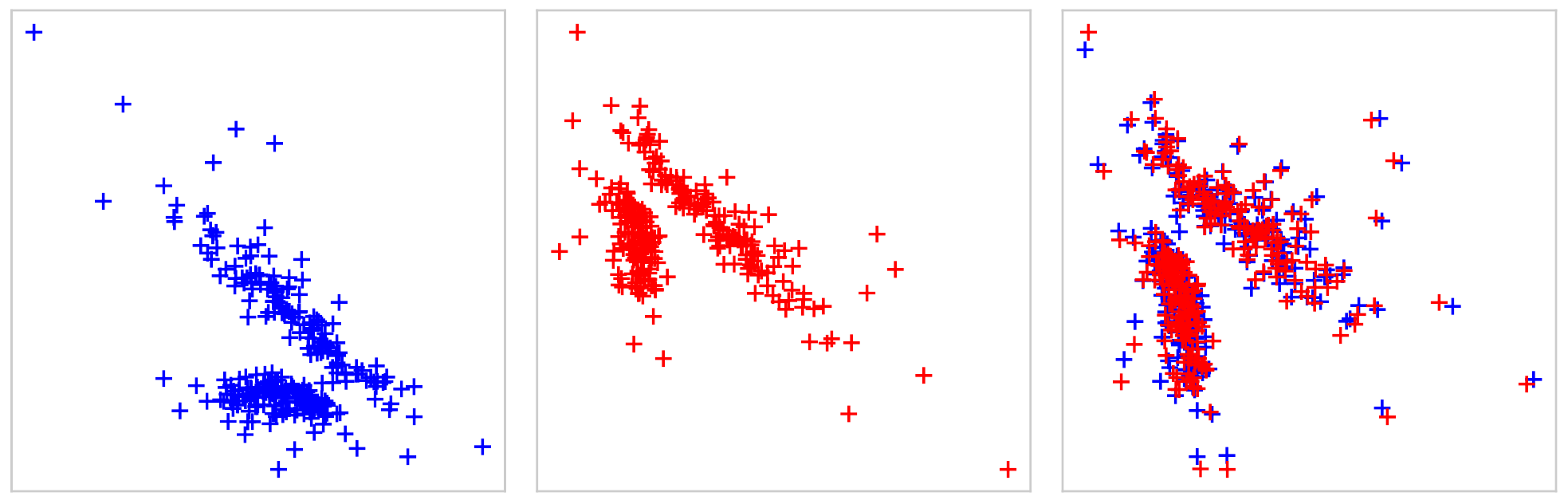}
  \end{center}
  \vspace{-1em}
  \caption{Two distinct 2-D representation functions parameterized by deep neural networks ${\color{blue} \bm{f}_{\bT_1}(\mathcal{B})}$, ${\color{red} \bm{f}_{\bT_2}(\mathcal{B})}$ on a subset $\mathcal{B}$ of validation dataset $\mathcal{D}$. We train on LM1B \citep{chelba2013one} using a word embedding model from \citet{mnih2012fast} (see Appendix \ref{app:fig1} for training details and code URL). The rightmost pane shows ${\color{blue} \bm{A}  \bm{f}_{\bT_1}(\mathcal{B})}$ and ${\color{red} \bm{f}_{\bT_2}(\mathcal{B})}$, where ${\color{blue} \bm{A} }$ is a linear transformation learned after training. This model exhibits {\it linear identifiability} (see Section \ref{sec:identifiability}): different representation functions, learned on the same data distribution, live within linear transformations of each other.}
\label{fig:fig1}
\vspace{-1em}
\end{figure}

\label{sec:gdl}
In this section, we define a discriminative model family for which we derive identifiability results in Section \ref{sec:identifiability}. Many discriminative models used for representation learning are members of this family, as we show in Section \ref{sec:lit}.

\subsection{Data Model}\label{sec:data}

We assume the existence of a generalized dataset in the form of an empirical distribution $\pD(\rvx, \rvy, \bS)$ over random variables $\rvx$, $\rvy$ and $\bS$ with the following properties:
\begin{itemize}
    \setlength\itemsep{0.0em}
    \item The random variable $\rvx$ is an input variable, typically high-dimensional such as text or an image.
    \item The random variable $\rvy$ is a target variable whose value the model predicts. In case of object classification, this would be some semantically meaningful class label. However, in our model family, $\rvy$ can also be a high-dimensional variable, such a text, image, or sentence fragment.
    \item $\bS$ is a set containing the possible values of $\rvy$ given $\rvx$, so $\pD(\rvy | \rvx, \bS) > 0 \iff \rvy \in \bS$. 
\end{itemize}
Note that unlike classical probabilistic discriminative learning, we treat $\bS$ as stochastic. 
We do so here in order to analyze common behaviour in learned representations among supervised, contrastive, and self-supervised learning frameworks that share the canonical mathematical form of equation \ref{eq:model}.
Briefly, when a model is supervised classification, $\bS$ is deterministic and contains class labels. When a model is self-supervised pretraining, the set $\bS$ is stochastic and contains high-dimensional variables such as image embeddings. When a model is deep metric learning \citep{hoffer2015deep, sohn2016improved}, the set $\bS$ is either and contains one positive and a random sample of negative examples of the group $\rvx$ belongs to. 

\subsection{Canonical Discriminative Form}\label{sec:model}
The model family is defined by its parametric probability of target context variable $\rvy$ given observed data $\rvx$ and a set $\bS$ comprising the true label $\rvy$ and a collection of distractors $\rvy^\prime$:
\begin{align}
    \pT(\rvy | \rvx, \bS) &= \frac{\exp(\rvfT(\rvx)^\top \rvgT(\rvy))}
    {\sum_{\rvy' \in \bS} \exp (\rvfT(\rvx)^\top \rvgT(\rvy'))},
\label{eq:model}
\end{align}
e.g., a softmax parameterization of a categorical log-likelihood with inner-product link function\footnote{Note that a bias parameter is accounted for by appending a constant 1 to $\rvfT$ and the bias to $\rvgT$.}.
The codomain of the functions $\rvfT(\rvx)$ and $\rvgT(\rvy)$ is  $\mathbb{R}^M$.
The domains vary with problem and modelling context.
For notational convenience both are parameterized by $\bT \in \Theta$, but $\rvf$ and $\rvg$ may use disjoint parts of $\bT$, such that they may or may not directly share parameters.
See Section \ref{sec:lit} and Appendix~\ref{app:gdl_equations} for extensive examples.

With $\mathcal{F}$ and $\mathcal{G}$ we denote the function spaces of functions $\rvfT$ and $\rvgT$ respectively.
Our primary domain of interest is when $\rvfT$ and $\rvgT$ are highly flexible function approximators, in particular deep neural networks.
In neural networks, different choices of parameters $\bT$ can result in the same functions $\rvfT$ and $\rvgT$, hence the map $\Theta \rightarrow \mathcal{F} \times \mathcal{G}$ is many-to-one. In the context of representation learning, the function $\rvfT$ is typically viewed as a nonlinear feature extractor, e.g., the learned representation of the input data.
While other choices meet the membership conditions for the family defined by the canonical form of Equation \ref{eq:model}, in the remainder, we will focus on deep neural networks.
The proof in Section \ref{sec:idanalysis} elaborates additional assumptions on $\rvfT$ and $\rvgT$ needed for identifiability.

\section{Model Identifiability}
\label{sec:identifiability}
In this section, we derive the identifiability conditions of models in the family defined in Section~\ref{sec:gdl}.
In short, sufficient conditions for identifiability by weakening the classical definition to admit identifiability up to a subset of the parameter space $\Theta$.

\subsection{Extending Identifiability to Learned Representations}
Prior to \citet{hyvarinen2016unsupervised}, identifiability analysis was uncommon in deep learning.
Identifiability analysis answers the question of whether it is theoretically possible to learn the parameters of a statistical model exactly.
Specifically, given some estimator $\bT'$ for model parameters $\bT^*$, identifiability is the property that, for any $\{ \bT', \bT^* \} \subset \Theta$, it is true that:
\begin{align}
     p_{\bT'} = p_{\bT^*} \;\; \implies \;\; \bT' = \bT^*.
\label{eq:idcl}
\end{align}
Models that do not have this property are said to be non-identifiable. This happens when different values $\{\bT', \bT^*\} \subset \Theta$ can give rise to the same model distribution $p_{\bT'}(\rvy | \rvx, \bS) = p_{\bT^*}(\rvy | \rvx, \bS)$. In such a case, observing an empirical distribution $p_{\bT^*}(\rvy | \rvx, \bS)$, and fitting a model $p_{\bT'}(\rvy | \rvx, \bS)$ to it perfectly, does not guarantee that $\bT' = \bT^*$. 

Neural networks exhibit various symmetries in parameter space such that there is almost always a many-to-one correspondence between a choice of $\bT$ and resulting probability function $\pT$. A simple example in neural networks is that one can swap the (incoming and outgoing) connections of two neurons in a hidden layer. This changes the value of the parameters, but does not change the network's function. Thus, when representation functions $\rvfT$ or $\rvgT$ are parameterized as deep neural networks, equation (\ref{eq:idcl}) is not reasonably satisfiable. Nevertheless, identifiability is a high-utility model property: if a model were identifiable, we could both guarantee that the representation functions learned were optimal for reproducibility and reliability reasons, as well as avoid resource-wasteful bespoke retraining. 
In the remainder, we derive identifiability conditions for the general class of discriminative models formalized in Section \ref{sec:gdl}, and explore their estimation through careful simulation and large-scale deep learning experiments in Section \ref{sec:experiments}.

\subsection{Identifiability Analysis}
\label{sec:idanalysis}
For reliable and efficient representation learning, we want learned representations $\rvfT$ from two identifiable models to be {\it sufficiently} similar for interchangeable use in downstream tasks.
The most general property we wish to preserve among learned representations is their ability to discriminate among statistical patterns among class or category groupings.
In the model family defined in Section \ref{sec:gdl}, the data and context functions $\rvfT$ and $\rvgT$ parameterizes $\pT(\rvy | \rvx, \bS)$, e.g., probability of group membership, through a normalized inner product.
This induces a hyperplane boundary for discrimination, in a joint space of learned representations for data $\rvx$ and context $\rvy$.
Therefore, in the following, we will derive identifiability conditions {\it up to a linear transformation}, using a notion of similarity in parameter space inspired by \citet{hyvarinen2018nonlinear}.

\definition{\it Let $\linsim$ be a pairwise relation on $\Theta$ 
defined as:
\label{prop:functional}
\begin{align}
    \bT' \linsim \bT^* \iff {
\rvf^\prime(\rvx) = {\bm A} \rvf^*(\rvx)
\atop
\rvg^\prime(\rvy) = {\bm B}  \rvg^*(\rvy)
}
\end{align}
where ${\bm A}$ and ${\bm B}$ are invertible $M \times M$ matrices. See Appendix \ref{app:def1} for proof that $\linsim$ is an equivalence relation.
}

In the remainder, we refer to identifiability up to the equivalence relation $\linsim$ as {\it $\linsim$-identifiable} or {\it linearly identifiable}.

We next present a simple derivation of the $\linsim$-identifiability of members of the generalized discriminative family defined in Section {\ref{sec:gdl}}.
The proof reveals sufficient conditions that deep neural networks, parameterizing $\rvfT$ and $\rvgT$, must satisfy for the model to be $\linsim$-identifiable. For brevity, we will use shorthands $p^\prime := p_{\bT^\prime}, p^* := p_{\bT^*}$, $\rvf^* := \rvf_{\bT^*}, \rvf^\prime := \rvf_{\bT^\prime}, \rvg^\prime := \rvg_{\bT^\prime}$, and drop superscripts when the statement applies to both.

We note an assumption on the data distribution and functions $\rvfT, \rvgT$. We assume that for any $(\bT', \bT^*)$ for which it holds that $p' = p^*$, and for any given $\rvx$, we assume that by repeated sampling $\bS \sim \pD(\bS | \rvx)$ and picking $\rvy_A, \rvy_B \in \bS$, we can construct a set of $M$ distinct tuples $\{(\rvy^{(i)}_A, \rvy^{(i)}_B)\}_{i=1}^M$ such that the matrices $\bb{L}'$ and $\bb{L}^*$ are invertible, where $\bb{L}'$ consists of columns $(\rvg'(\rvy^{(i)}_A) - \rvgT(\rvy^{(i)}_B))$, and $\bb{L}^*$ consists of columns $(\rvg^*(\rvy^{(i)}_A) - \rvg^*(\rvy^{(i)}_B))$, with $i = 1, \dots, M$. We refer to this as the {\it diversity condition}; see Section~\ref{sec:assumptions} for discussion. 

\theorem{ \label{th:1}Under the diversity condition above, models in the family defined by Equation (\ref{eq:model}) are  linearly identifiable. That is, for any $\bT^\prime, \bT^* \in \Theta$, and $\rvf^*, \rvf^\prime, \rvg^*, \rvg^\prime, p^*, p'$ defined as in Section {\ref{sec:gdl}},
\begin{align}
     p' = p^{*} \;\; \implies \;\; \bT' \linsim \bT^*.
\label{eq:linid}
\end{align}

\noindent Please note that the straightforward proof in this section builds on similar theoretical results presented in earlier papers. Please see Section \ref {sec:related_work} for a detailed comparison.
}
\paragraph{Proof.} 
We proceed by constructing the invertible linear transformations to satisfy Definition~\ref{prop:functional}. Consider $\rvy_A, \rvy_B  \in \bS$. The likelihood ratios for these points
\begin{align}
    \frac{p'(\rvy_A|\rvx, \bS)}{p'(\rvy_B|\rvx, \bS)}
    = \frac{p^*(\rvy_A|\rvx, \bS)}{p^*(\rvy_B|\rvx, \bS)}
\end{align}
are equal. Substituting our model definition from equation (\ref{eq:model}), we find:
    \begin{align}
        \frac{\exp( \rvf'(\rvx)^\top \rvg'(\rvy_A))}{\exp (\rvf'(\rvx)^\top \rvg'(\rvy_B) )}
        =
        \frac{\exp( \rvf^*(\rvx)^\top \rvg^*(\rvy_A))}{\exp (\rvf^*(\rvx)^\top \rvg^*(\rvy_B) )},
        \label{eq:normalizing}
\end{align}
where we note that the normalizing constants $\sum_{\rvy \in \bS} \exp (\rvf(\rvx)^\top \rvg(\rvy))$ cancelled out on the left- and right-hand sides. Taking the logarithm of both sides, this simplifies to:
\begin{align}
    & (\rvg'(\rvy_A) - \rvg'(\rvy_B))^\top \rvf'(\rvx)
    = (\rvg^*(\rvy_A) - \rvg^*(\rvy_B) )^\top \rvf^*(\rvx).
\label{eq:simple}
\end{align}

Note that this equation is true for any triple $(\rvx, \rvy_{\tiny A}, \rvy_B)$ for which $\pD(\rvx, \rvy_B, \rvy_B ) > 0$.



We next collect $M$ distinct tuples $(\rvy^{(i)}_A, \rvy^{(i)}_B)$ so that by repeating Equation (\ref{eq:simple}) $M$ times and by the diversity condition noted above, the resulting difference vectors are linearly independent. We collect these vectors together as the columns of $(M \times M)$-dimensional matrices $\bb{L}'$ and $\bb{L}^*$, forming the following system of $M$ linear equations:
\begin{align*}
    \bb{L}'^\top \rvf'(\rvx) &= \bb{L}^{*\top} \rvf^*(\rvx).
\end{align*}
Since $\bb{L}'$ and $\bb{L}^*$ are invertible, we rearrange:
\begin{align}
    \rvf'(\rvx) &= (\bb{L}^{*} \bb{L}'^{-1})^\top\, \rvf^*(\rvx).
\label{eq:all}\end{align}
Therefore, $\rvf^\prime(\rvx) = {\bb A} \rvf^*(\rvx)$ where $\bb{A} = (\bb{L}^{*} \bb{L}'^{-1})$ is invertible. In Appendix \ref{sec:identifiability_g} we give a similar proof for linear identifiability of $\rvgT$. \\ \qed

\subsection{Discussion of Diversity Condition}
\label{sec:assumptions}

Theorem \ref{th:1} requires invertible $(M \times M)$ matrices $\bf{L}'$ and $\bb{L}^*$. This requirement is similar to the conditions in earlier work on nonlinear ICA such as ~\citep{hyvarinen2018nonlinear}, as discussed in Section \ref{sec:related_work}. 
Informally, this means that there needs to be a sufficient number of possible values $\rvy \in \bS$.
In the case of a supervised classification with $K$ classes, $\bS$ is fixed and of size $K$.
Then, we need $K \geq M+1$ in order to generate $M$ difference vectors $\rvgT(\rvy^{(1)}) - \rvgT(\rvy^{(j)})$, $j = 2, \dots, M+1$.
 In case of self-supervised or deep metric learning, where $\bS$ and $\rvy$ may be stochastically generated from $\rvx$, there will typically be a diversity of values of $\rvy$.
 
Note that by the diversity requirement, we implicitly assume $\rvgT$ to have the following property: the $M$ difference vectors span the range of $\rvgT$.
This is a mild assumption in the context of deep neural networks: for random initialization and iterative weight updates, this property follows from the sampling distribution used to choose the initial weights.
Briefly, a set of $M+1$ unique points $\rvy^{(j)}$ such that the $M$ vectors $\rvgT(\rvy^{(1)}) - \rvgT(\rvy^{(j)}), j=2, \dots, M+1$ are not linearly independent has measure zero.
For other choices of $\rvgT$, care must be taken to ensure this condition is satisfied.

\section{Varieties of Linearly Identifiable Models}
\label{sec:lit}

The conditions for Theorem \ref{th:1} hold in a variety of models for different problem contexts, often state-of-the-art at time of publication: Contrastive Predictive Coding \citep{henaff2019data}, BERT \citep{devlin2018bert}, GPT-2 and GPT-3 \citep{radford2018improving, radford2019language, brown2020language}, XLNET \citep{yang2019xlnet}, and the triplet loss for deep metric learning \citep{sohn2016improved}.
Functions $\rvfT$ and $\rvgT$ have been implemented using a variety of deep learning architectures.
We discuss these here.
For reductions to the canonical form of Equation~\ref{eq:model}, please refer to Appendix {\ref{app:gdl_equations}}.

\paragraph{Deep Supervised Classification.} 
Classification models that deploy deep neural networks as feature extractors satisfy the sufficient conditions given in Section \ref{sec:identifiability} when the following is true: the network from input to the layer prior to the logits is the representation function $\rvfT(\rvx)$, and weights in the final projection layer are given by the context map that depends only on the labels. The class labels $\rvy$ are integers $0$ to $K-1$, and $\rvw_i = \rvgT(\rvy = i)$ is the $i$-th column of a weight matrix $\bb{W}$, and the vector representation of the $i$-th class.
The set $\bS$ in this case is a constant, containing semantically meaningful labels. Other than a careful simulation study to empirically validate the above claim (Section \ref{sec:simulation}), we do not explore supervised classification further. Self-supervised and multi-task pretraining have been empirically shown to be more data efficient for learning effective representations. We leave an empirical analysis of efficient estimation for future work.

\paragraph{{Self-Supervised Pretraining for Image Classification}.}
Self-supervised learning is supervised learning with algorithmically-generated labels.
Typically, self-supervision is used to pretrain network weights using a classification task with synthetically-generated data and labels, in order to improve performance a downstream, related task. Contrastive Predictive Coding (CPC) \citep{oord2018representation} is a paradigmatic self-supervised method, often applied to image domains, and is a member of the model family of Section \ref{sec:gdl}. 
CPC as applied to images involves: (1) pre-processing the image into patches, (2) assigning labels according to which image the patch came from, and then (3) predicting the representations of the patches whether below, to the right, to the left, or above a certain level \citep{oord2018representation}. 

The context function of CPC, $\rvgT(\rvy)$, encodes a particular position in the sequence of patches, and the representation function, $\rvfT(\rvx)$, is an autoregressive function of the previous $k$ patches, according to some predefined patch ordering.
The collection of all patches from the sequence, from a given minibatch of images, the set $\bS$. 

\paragraph{Multi-task Pretraining for Natural Language Generation.} Autoregressive language models, such as \citep{mikolov2010recurrent, dai2015semi} and more recently GPT-2 and GPT-3 \citep{radford2018improving, radford2019language, brown2020language}, are typically also instances of the model family of \Eqref{eq:model}. 
Data points $\rvx$ are the past tokens, $\rvfT(\rvx)$ is a nonlinear representation of the past estimated by either an LSTM~\citep{hochreiter1997long} or an autoregessive Transformer model~\citep{vaswani2017attention}, $\rvy$ is the next token, and $\rvw_i = \rvgT(\rvy = i)$ is a learned representation of the next token, typically implemented as a simple look-up table. 

BERT \citep{devlin2018bert} learns word embeddings through a denoising autoencoder-like \citep{vincent2008extracting} architecture. 
For a given sequence of tokenized text, some fixed percentage of the symbols are extracted and set aside, and their original values set to a special null symbol, ``corrupting" the original sequence.
The pre-training task in BERT is to learn a continuous representation of the extracted symbols conditioned on the remainder of the text.
A transformer \citep{vaswani2017attention} function approximator is used to map from the corrupted sequence into a continuous space.
The transformer network is the $\rvfT(\rvx)$ function of \Eqref{eq:model}. 
The context map $\rvgT(\rvy)$ is a lookup map into the unlearned basis vector for each token.

\section{Experimental Validation}
\label{sec:experiments}

The derivation in Section \ref{sec:identifiability} shows that, for models in the general discriminative family defined in Section \ref{sec:gdl}, the functions $\rvfT$ and $\rvgT$ are asymptotically identifiable up to a linear transformation, e.g., given unbounded data and model convergence.
The question remains as to how close a model trained on finite data and without convergence guarantees will approach this limit--that is, the domain of deep learning practice.
Results in this section present evidence for: (1) close convergence in the small dimensional, large data regime; and (2) monotonic increase in linear similarity of learned representations as a function of dataset size and model capacity in the high dimensional regime.
Note that due to the generality of $\rvfT$ and $\rvgT$ and Theorem $\ref{th:1}$, the number of possible experiments is huge.
Here, we focus on a core set of linearly identifiable models, interpolating from a low dimensional simulation study of deep supervised classification to GPT-2 \citep{radford2019language}, an approximately $1.5*10^9$-parameter generative model of natural language.
For brevity, we have suppressed fine-grained training and model details.
Please see Appendix \ref{app:train} for additional details needed to reproduce.

\subsection{Measuring linear similarity between learned representations}
 Our goal is to {\it measure} whether pairs of learned representations live within a linear transformation of each other.
We adapt Canonical Correlation Analysis (CCA) \citep{hotelling1936relations} for this purpose, which finds the optimal linear transformations to maximize correlation among two random vectors.
We select a subset $\mathcal{B} \subset \mathcal{D}$ of the training data and compute $\bm{f}_{\bT_1}(\mathcal{B})$ and $\bm{f}_{\bT_2}(\mathcal{B})$, for two models with parameters $\bT_1$ and $\bT_2$ respectively.
Assume without loss of generality that $\bm{f}_{\bT_1}(\mathcal{B})$ and $\bm{f}_{\bT_2}(\mathcal{B})$ are centered.
Then, CCA finds the optimal linear transformations $\bm{C}$ and $\bm{D}$ such that the pairwise correlations $\rho_i = \operatorname{Corr}[\bm{C}_i^\top \bm{f}_{\bT_1}(\mathcal{B}), \bm{D}_i^\top \bm{f}_{\bT_2}(\mathcal{B})]$ are maximized.
If one vector is a linear transform of another, CCA applied to both will learn this transformation and the mean of $\boldsymbol{\rho}$ will be 1; if they are perfectly uncorrelated then the mean of $\boldsymbol{\rho}$ is 0.
We use the mean of $\boldsymbol{\rho}$ as a proxy for the existence of a linear transformation between $\bm{f}_{\bT_1}(\mathcal{B})$ and $\bm{f}_{\bT_2}(\mathcal{B})$. 

For deep neural networks, it is a well known phenomenon that most of the variability in a learned representation tends to concentrate in a low-dimensional subspace, leaving many noisy, random dimensions \citep{morcos2018insights}.
Such random noise can result in spurious high correlations in CCA.
A solution to this problem is to apply Principal Components Analysis (PCA) \citep{pearson1901liii} to each of the two images $\bm{f}_{\bT_2}(\mathcal{B})$ and $\bm{f}_{\bT_1}(\mathcal{B})$, projecting onto their top-$k$ principal components, before applying CCA.
This technique is known as SVCCA \citep{raghu2017svcca}.

\subsection{Simulation Study: Classification by Deep Neural Networks}
\label{sec:simulation}

\begin{figure*}[th]
\centering
\begin{overpic}[width=\textwidth]{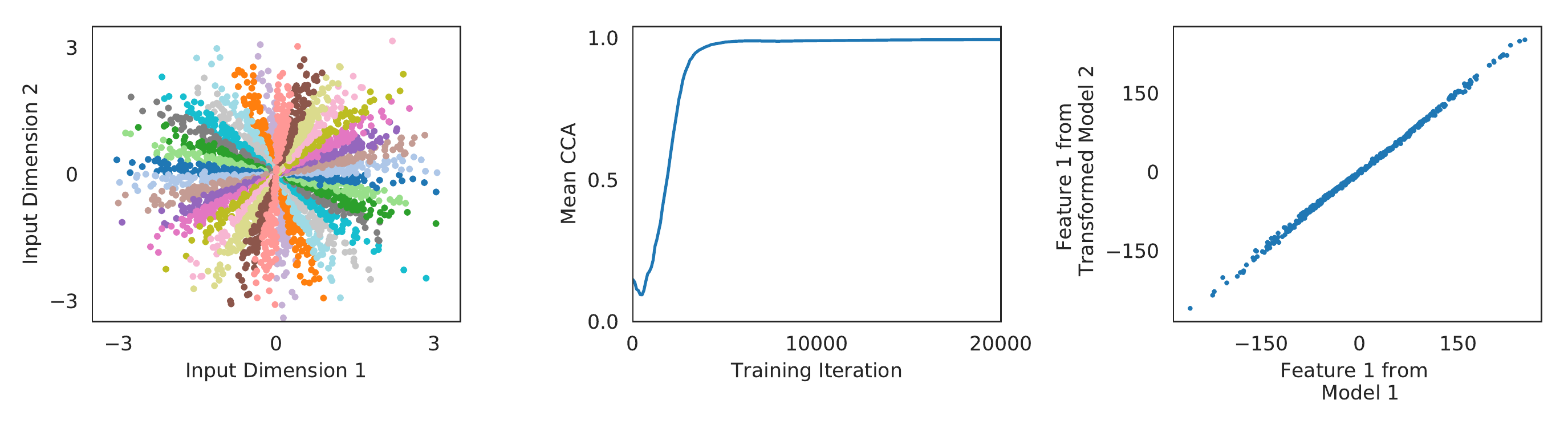}
 \put (0,2.0) {\textbf{\small(a)}}
 \put (35,2.0) {\textbf{\small(b)}}
 \put (69,2.0) {\textbf{\small(c)}}
\end{overpic}
  \vspace{-1em}
  \caption{{\bf Deep Supervised Classification}. \textbf{(a)} Data distribution for a linearly identifiable K-way classification problem. 
  \textbf{(b)} Mean (centered) CCA between the learned representations over the course of training. After approx. 4000 iterations, CCA finds a linear transformation that rotate the learned representations into alignment, up to optimization error.
  \textbf{(c)} Learned representations after transformation via optimal linear transformation. The first dimension of the first model's feature space is plotted against the first dimension of second. The learned representations have a nearly linear relationship, modulo estimation noise.
  \label{fig:synth}}
\end{figure*}

We conducted simulation study of linearly identifiable $K$-way classification, where all assumptions and sufficient conditions are met by construction.

First, we designed a data distribution with the properties required by Section \ref{sec:gdl}.
The data distribution $\pD(\rvx, \rvy, \bS)$ was generated as follows: points $\rvx$ are sampled from a 2-D Gaussian with $\sigma = 3$ and assigned pre-labels among $K$ classes radially, for $K=18$ ($\sigma$ chosen arbitrarily and $K$ to ensure $K \geq M+1$).
The context map is $\rvgT(\rvy) = \bb{W} \rve_{\rvy}$ where $\rve_{\rvy}$ is a one-hot encoding of the class identity.

For this experiment, we want to minimize noise in the model parameters due to model misspecification.
Model misspecification would add noise optimizing the parameters of the function approximators $\rvf_{\bT^\prime}$ and $\rvg_{\bT^\prime}$ to $\rvf_{\bT^\star}$ and $\rvg_{\bT^\star}$. 
We eliminated model misspecification by fitting two neural networks $f_{\bT^\star}$ and $g_{\bT^\star}$ to predict the labels according to Equation (\ref{eq:model}).
In particular, both are 4-hidden-layer MLPs with two 64 unit layers and one 2-D bottle neck layer.
We then used these representation functions to predict ground truth labels for each $\rvx$.
Finally, we chose $\rvf_{\bT^\prime}$ and $\rvg_{\bT^\prime}$ to be the same architecture as $\rvf_{\bT^\star}$ and $\rvg_{\bT^\star}$, ensuring that the parameters $\bT^\prime$ and $\bT^\star$ live in the same space $\Theta$.
This enforces that the function approximators $\rvf_{\bT^\prime}$ and  $\rvg_{\bT^\prime}$ are able to represent the true data-generating process.

To evaluate linear similarity, we trained two randomly initialized models of $\pT(\rvy | \rvx, \bS)$ with the same architecture as the data generating process.
We visualize the image of $\rvfT$, the data representation function.
Figure \ref{fig:synth}b shows that the mean CCA increases to its maximum value over training, demonstrating that the feature spaces converge to the same solution up to a linear transformation up to estimation noise.
Similarly, Figure \ref{fig:synth}c shows that the learned representations, after mapping through the optimal linear transformation from CCA, have a strongly linear relationship.

\subsection{Self-Supervised Learning for Image Classification}
\begin{figure*}[th]
\centering
\begin{overpic}[width=\textwidth]{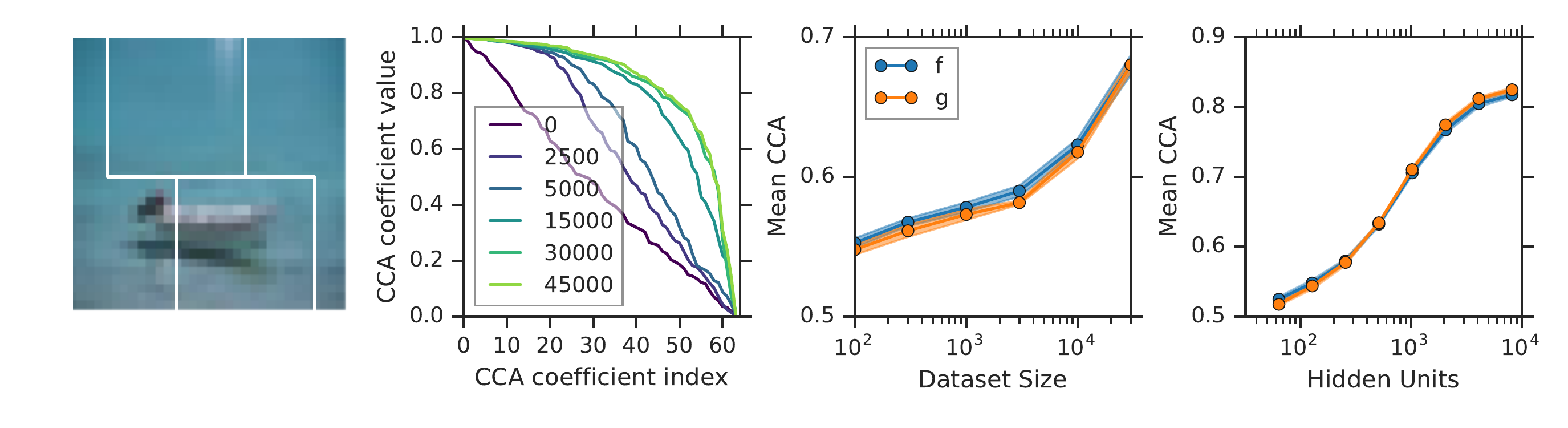}
 \put (1,3.0) {\textbf{\small(a)}}
 \put (25,3.0) {\textbf{\small(b)}}
 \put (50,3.0) {\textbf{\small(c)}}
 \put (75,3.0) {\textbf{\small(d)}}
\end{overpic}
  \vspace{-2.5em}
  \caption{{\bf Self-Supervised Representation Learning.}~\textbf{(a)} Input data. Two patches are taken (one from top half, and one from the bottom half) of an image at random. Using a contrastive loss, we predict the identity of the bottom patch encoding from the top patch encoding.
  \textbf{(b)} Linear similarity of learned representations at checkpoints (see legend). As models converge, linear similarity increases.
  \textbf{(c)} 
  Linear similarity as we increase the amount of data for $\rvfT$ and $\rvgT$. Error bars are computed over 5 pairs of models.
  \textbf{(d)} As we increase model size, linear similarity after convergence increases for both $\rvfT$ and $\rvgT$. Error bars are computed over 5 pairs of models.
  \label{fig:cpc_cifar}
  }
\end{figure*}

\begin{figure*}[th]
\centering
  \includegraphics[width=\textwidth]{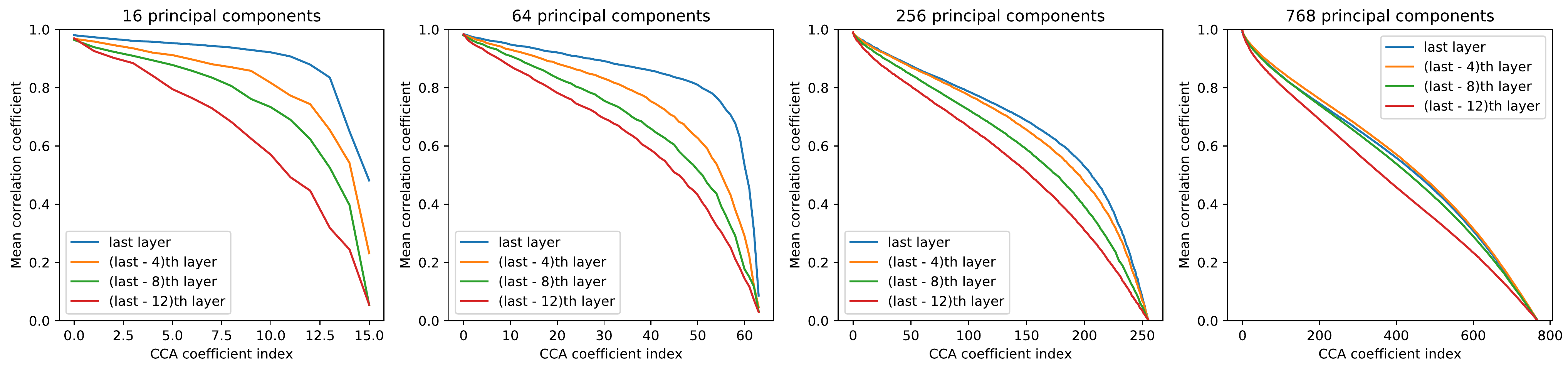}
  \vspace{-1.5em}
  \caption{{\bf Word Embeddings by GPT-2}. GPT-2 results. We computed representations of the last hidden layer (which is identifiable), in addition to three earlier layers (not necesarilly identifiable) for four GPT-2 models \citep{radford2019language}. For each representation layer, we then applied SVCCA to all possible pairs of models and averaged their correlation coefficients. SVCCA was applied with 16, 64, 256 and 768 principal components. The results show that, in line with expectations, the learned representations in the \emph{last} layer are much more correlated between the four models than the representations learned in preceding layers.
  } \vspace{-.5em}
    \label{fig:gpt2_layers}

\end{figure*}
We next investigate Theorem $\ref{th:1}$ when we can no longer expect efficient model estimation or nice distributional properties: high-dimensional, self-supervised representation learning on CIFAR-10 \citep{krizhevsky2009learning} using CPC \citep{oord2018representation,  henaff2019data}.
For a given input image, this model predicts the identity of a bottom image patch representation given a top patch representation (Figure \ref{fig:cpc_cifar}a.)
Here, $\bS$ comprises a the true patch with a set of distractor patches from across the minibatch.
For each model we define both $\rvf_{\bT^\prime}$ and $\rvg_{\bT^\prime}$ as a 3-layer MLP with 256 units per layer (except where noted otherwise) and fix output dimensionality of 64.

In Figure \ref{fig:cpc_cifar}b, we plot CCA coefficients over the course of training.
As training progresses, the linear similarity between the learned representations increases.
In Figure \ref{fig:cpc_cifar}c, we artificially limit the size of the dataset, and plot mean correlation after retraining and convergence.
This shows that increasing availability of data correlates with closer linear similarity.
In Figure \ref{fig:cpc_cifar}d, we fix dataset size and artificially limit the model capacity (number hidden units) to investigate the effect of model size on the learned representations, varying the number of hidden units from 64 to 8192.
This show that increasing model size correlates strongly with increase in linear similarity of learned representations.
Despite lack of estimation guarantees, these experiments validate Theorem $\ref{th:1}$.

\subsection{GPT-2}

Next, we performed experiments with GPT-2~\citep{radford2019language}, a large-scale language model. The identifiable representation is, as usual, the set of features just before the last linear layer of the model. 
We use pre-trained models from HuggingFace
\citep{Wolf2019HuggingFacesTS}.
This repository provides four different versions of the GPT-2: \texttt{gpt2}, \texttt{gpt2-medium}, \texttt{gpt2-large} and \texttt{gpt2-xl}, which differ mainly in the hyper-parameters that determine the width and depth of the neural network layers.
For approximately 2000 input sentences, per timestep, for each model, we extracted representations at the last layer (which is identifiable) in addition to the representations per timestep given by three earlier layers in the model. 
Then, we performed SVCCA on each possible pair of models, on each of the four representations. 
SVCCA was performed with 16, 64, 256 and 768 principal components, computed by applying SVD separately for each representations of each model. 
We chose 768 as the largest number of principal components, since that is the representation size for the smallest model in the repository (\texttt{gpt2}). 
We then averaged the CCA correlation coefficients across the pairs of models.

Figure~\ref{fig:gpt2_layers} shows the results. The results align well with our theory, namely that the representations at the last layer are more linearly related than the representations at other layers of the model.

\section{Related Work}
\label{sec:related_work}
This work extends early results presented at \citep{roeder2019identifiability}. We build on advances in the theory of nonlinear ICA~\citep{hyvarinen2016unsupervised,hyvarinen2018nonlinear,khemakhem2019variational}. Our diversity assumption is similar to diversity assumptions in these earlier works, while differing in certain conditions. The main difference with our result is these earlier works differ in their model families, such that they apply to related but distinct families of models than the general discriminative family outlined in this paper. Our proofs are similar to these earlier works, albeit simpler, due to a lack of an explicit probabilistic generative model.

Arguably most relevant is Theorem 3 of ~\citep{hyvarinen2018nonlinear} and its proof, which shows that a class of contrastive discriminative models will estimate, up to an affine transformation, the true latent variables of a nonlinear ICA model. The main difference with our result is that they additionally assume: (1) that the mapping between observed variables and latent representations is invertible; and (2) that the discriminative model is binary logistic regression exhibiting universal approximation \citep{hornik1989multilayer}, estimated with a contrastive objective. In addition, ~\citep{hyvarinen2018nonlinear} does not present conditions for affine identifiability for their version of the context representation function $\rvg$. It should be noted that Theorem 1 in ~\citep{hyvarinen2018nonlinear} provides a potential avenue for further generalization of our theorem \ref{th:1} to discriminative models with non-linear interaction between $\rvf$ and $\rvg$.

Concurrent work \citep{khemakhem2020ice} has expanded the theory of identifiable nonlinear ICA to a class of conditional energy-based models (EBMs) with universal density approximation capability, therefore imposing milder assumptions than previous nonlinear ICA results. Their version of affine identifiability is similar to our result of linear identifiability in Section \ref{sec:idanalysis}. The main differences are that \citet{khemakhem2020ice} focus in both theory and experiment on EBMs. This allows for alternative versions of the diversity condition, assuming that the Jacobians of their versions of $\rvf$ or $\rvg$ are full rank. This is only possible if $\rvx$ or $\rvy$ are assumed continuous-valued; note that we do not make such an assumption. \citep{khemakhem2020ice} also presents an architecture for which the conditions provably hold, in addition to sufficient conditions for identifiability up to element-wise scaling, which we did not explore in this work.

While we build on these earlier results, we are, to the best of our knowledge, the first to bridge the gap with state-of-the-art discriminative and autoregressive generative models.

\section{Conclusion}
We have shown that representations learned by a large family of discriminative models are identifiable up to a linear transformation, providing a new perspective on representation learning using deep neural networks.
Since identifiability is a property of a model class, and identification is realized in the asymptotic limit of data and compute, we perform experiments in the more realistic setting with finite datasets and finite compute. 
We demonstrate that as one increases the representational capacity of the model and dataset size, learned representations indeed tend towards solutions that are equal up to only a linear transformation.

\section{Acknowledgements}
We wish to acknowledge Ryan Adams for in-depth conversations on methodology, Kevin Murphy for helpful feedback on an early version, Aapo Hyv{\"a}rinen, Ilyes Khemakhem, and Jascha Sohl-Dickstein for helpful conversations and feedback, George Tucker and Alexi Alemi for feedback on an early version of the theorem, and members of the Google Brain Team and Princeton Laboratory for Intelligent Probabilistic Systems for valuable remarks and comments.

Geoffrey Roeder is supported in part by the National Science and Engineering Research Council of Canada (PGSD3-518716-2018).

\newpage
\section*{Broader Impact}

Identifiability results, like the one presented in this paper, are (as we aim to demonstrate) helpful in predicting when learned optimal representations are easily reproducible or not. While non-identifiable models might need to be trained many times to reproduce optimal results, identifiable architectures would (in principle) only need to be trained once, reducing the amount of computational resources required, thereby reducing waste, as well as saving budget and labour time.
Moreover, the field of deep learning at time of writing has been characterized by increasingly efficient function approximators trained on growing quantities of data. 
This trend interfaces with the asymptotic character of linear identifiability
in a pleasing way: as network architecture design continues to improve and more data is collected, we expect the representation functions approximated by deep neural networks to approach a stable set of optima in function space.
Our hope is that such an identifiable deep learning can ameliorate the environmentally unsustainable, massive expenditure of carbon-based and non-renewable energy resources, replacing it with a library of provably optimal trained models for particular tasks on particular datasets.

This possibility also highlights a notable risk of Theorem \ref{th:1}: models in the family of Section \ref{sec:gdl} may be trained to learn {\it biased} representations that are ``fixed'' for particular datasets and for particular tasks.
Learned representations in general obfuscate societal biases by making them appear as if they were a consequence of impartial statistical patterns, amplifying their negative impact on marginalized communities by granting a false veneer of certified objectivity.
Such biases are not impartial statistical patterns: they arise through the choice of discriminative task, collection of training and validation data, the architecture of the representation and context functions, and objective function itself \citep{denton2020data}.
Model cards \citep{mitchell2019model} are one way to expose potential biases along these dimensions at time of writing, and we strongly advise their use here.

On the other hand, identifiable representation learning presents an opportunity to produce optimal learned representations that the broader community agrees reflect its values and ethics. With such ethical representations, we as a scientific community can work to prevent tacit reproduction and thereby perpetuation of racism, sexism, and other harmful prejudices.

\bibliography{4_refs.bib}

\begin{thebibliography}{37}
\providecommand{\natexlab}[1]{#1}
\providecommand{\url}[1]{\texttt{#1}}
\expandafter\ifx\csname urlstyle\endcsname\relax
  \providecommand{\doi}[1]{doi: #1}\else
  \providecommand{\doi}{doi: \begingroup \urlstyle{rm}\Url}\fi

\bibitem[Bradbury et~al.(2018)Bradbury, Frostig, Hawkins, Johnson, Leary,
  Maclaurin, and Wanderman-milne]{jax2018github}
J.~Bradbury, R.~Frostig, P.~Hawkins, M.~J. Johnson, C.~Leary, D.~Maclaurin, and
  S.~Wanderman-milne.
\newblock {Jax}: Composable transformations of {P}ython+{N}um{P}y programs,
  2018.
\newblock URL \url{Http://Github.Com/Google/Jax}.

\bibitem[Brown et~al.(2020)Brown, Mann, Ryder, Subbiah, Kaplan, Dhariwal,
  Neelakantan, Shyam, Sastry, Askell, and Others]{brown2020language}
T.~B. Brown, B.~Mann, N.~Ryder, M.~Subbiah, J.~Kaplan, P.~Dhariwal,
  A.~Neelakantan, P.~Shyam, G.~Sastry, A.~Askell, and Others.
\newblock {Language Models are Few-Shot Learners}.
\newblock \emph{Arxiv Preprint Arxiv:2005.14165}, 2020.

\bibitem[Chelba et~al.(2013)Chelba, Mikolov, Schuster, Ge, Brants, Koehn, and
  Robinson]{chelba2013one}
C.~Chelba, T.~Mikolov, M.~Schuster, Q.~Ge, T.~Brants, P.~Koehn, and
  t.~Robinson.
\newblock {One Billion Word Benchmark for Measuring Progress in Statistical
  Language Modeling}.
\newblock \emph{Arxiv Preprint Arxiv:1312.3005}, 2013.

\bibitem[Dai and Le(2015)]{dai2015semi}
A.~M. Dai and Q.~V. Le.
\newblock {Semi-Supervised Sequence Learning}.
\newblock In \emph{Advances in Neural information Processing Systems}, pages
  3079--3087, 2015.

\bibitem[Denton and Gebru(2020)]{denton2020data}
E.~Denton and T.~Gebru.
\newblock {Data Ethics}.
\newblock Tutorial on Fairness Accountability Transparency and Ethics in
  Computer Vision at CVPR 2020, 2020.
\newblock URL
  \url{https://drive.google.com/file/d/1IvUgCTUciIJQ-dIqQAYNO11X3guzqnYN/view}.

\bibitem[Devlin et~al.(2018)Devlin, Chang, Lee, and Toutanova]{devlin2018bert}
J.~Devlin, M.-W. Chang, K.~Lee, and K.~Toutanova.
\newblock {BERT: Pre-training of Deep Bidirectional Transformers for Language
  Understanding}.
\newblock \emph{Arxiv Preprint Arxiv:1810.04805}, 2018.

\bibitem[Erhan et~al.(2010)Erhan, Bengio, Courville, Manzagol, Vincent, and
  Bengio]{erhan2010does}
D.~Erhan, Y.~Bengio, A.~Courville, P.-A. Manzagol, P.~Vincent, and S.~Bengio.
\newblock {Why Does Unsupervised Pre-training Help Deep Learning?}
\newblock \emph{Journal of Machine Learning Research}, 11\penalty0
  (Feb):\penalty0 625--660, 2010.

\bibitem[H{\'e}naff et~al.(2019)H{\'e}naff, Razavi, Doersch, Eslami, and
  Oord]{henaff2019data}
O.~J. H{\'e}naff, A.~Razavi, C.~Doersch, S.~Eslami, and A.~V.~D. Oord.
\newblock {Data-Efficient Image Recognition with Contrastive Predictive
  Coding}.
\newblock \emph{Arxiv Preprint Arxiv:1905.09272}, 2019.

\bibitem[Hochreiter and Schmidhuber(1997)]{hochreiter1997long}
S.~Hochreiter and J.~Schmidhuber.
\newblock {Long Short-Term Memory}.
\newblock \emph{Neural Computation}, 9\penalty0 (8):\penalty0 1735--1780, 1997.

\bibitem[Hoffer and Ailon(2015)]{hoffer2015deep}
E.~Hoffer and N.~Ailon.
\newblock {Deep Metric Learning Using Triplet Network}.
\newblock In \emph{International Workshop On Similarity-Based Pattern
  Recognition}, pages 84--92. Springer, 2015.

\bibitem[Hornik et~al.(1989)Hornik, Stinchcombe, and
  White]{hornik1989multilayer}
K.~Hornik, M.~Stinchcombe, and H.~White.
\newblock {Multilayer Feedforward Networks are Universal Approximators}.
\newblock \emph{Neural Networks}, 2\penalty0 (5):\penalty0 359--366, 1989.

\bibitem[Hotelling(1936)]{hotelling1936relations}
H.~Hotelling.
\newblock {Relations Between Two Sets of Variates}.
\newblock \emph{Biometrika}, 28\penalty0 (3/4):\penalty0 321--377, 1936.

\bibitem[Hyv{\"a}rinen and Morioka(2016)]{hyvarinen2016unsupervised}
A.~Hyv{\"a}rinen and H.~Morioka.
\newblock {Unsupervised Feature Extraction by Time-Contrastive Learning and
  Nonlinear ICA}.
\newblock In \emph{Advances in Neural information Processing Systems}, pages
  3765--3773, 2016.

\bibitem[Hyv{\"a}rinen et~al.(2018)Hyv{\"a}rinen, Sasaki, and
  Turner]{hyvarinen2018nonlinear}
A.~Hyv{\"a}rinen, H.~Sasaki, and R.~E. Turner.
\newblock {Nonlinear ICA Using Auxiliary Variables and Generalized Contrastive
  Learning}.
\newblock \emph{Arxiv Preprint Arxiv:1805.08651}, 2018.

\bibitem[Khemakhem et~al.(2019)Khemakhem, Kingma, and
  Hyv{\"a}rinen]{khemakhem2019variational}
I.~Khemakhem, D.~P. Kingma, and A.~Hyv{\"a}rinen.
\newblock {Variational Autoencoders and Nonlinear ICA: A Unifying Framework}.
\newblock \emph{Arxiv Preprint Arxiv:1907.04809}, 2019.

\bibitem[Khemakhem et~al.(2020)Khemakhem, Monti, Kingma, and
  Hyv{\"a}rinen]{khemakhem2020ice}
I.~Khemakhem, R.~P. Monti, D.~P. Kingma, and A.~Hyv{\"a}rinen.
\newblock {ICE-BeeM: Identifiable Conditional Energy-based Deep Models}.
\newblock \emph{Arxiv Preprint Arxiv:2002.11537}, 2020.

\bibitem[Krizhevsky et~al.(2009)Krizhevsky, Hinton, and
  Others]{krizhevsky2009learning}
A.~Krizhevsky, G.~Hinton, and Others.
\newblock {Learning Multiple Layers of Features from Tiny Images}.
\newblock 2009.

\bibitem[Liu et~al.(2018)Liu, Saleh, Pot, Goodrich, Sepassi, Kaiser, and
  Shazeer]{liu2018generating}
P.~J. Liu, M.~Saleh, E.~Pot, B.~Goodrich, R.~Sepassi, L.~Kaiser, and
  N.~Shazeer.
\newblock {Generating Wikipedia by Summarizing Long Sequences}.
\newblock \emph{Arxiv Preprint Arxiv:1801.10198}, 2018.

\bibitem[Mikolov et~al.(2010)Mikolov, Karafi{\'a}t, Burget, {\v{C}}ernock{\`y},
  and Khudanpur]{mikolov2010recurrent}
T.~Mikolov, M.~Karafi{\'a}t, L.~Burget, J.~{\v{C}}ernock{\`y}, and
  S.~Khudanpur.
\newblock {Recurrent Neural Network Based Language Model}.
\newblock In \emph{Eleventh Annual Conference of The international Speech
  Communication Association}, 2010.

\bibitem[Mikolov et~al.(2013)Mikolov, Sutskever, Chen, Corrado, and
  Dean]{mikolov2013distributed}
T.~Mikolov, I.~Sutskever, K.~Chen, G.~S. Corrado, and J.~Dean.
\newblock {Distributed Representations of Words and Phrases and their
  Compositionality}.
\newblock In \emph{Advances in Neural information Processing Systems}, pages
  3111--3119, 2013.

\bibitem[Mitchell et~al.(2019)Mitchell, Wu, Zaldivar, Barnes, Vasserman,
  Hutchinson, Spitzer, Raji, and Gebru]{mitchell2019model}
M.~Mitchell, S.~Wu, A.~Zaldivar, P.~Barnes, L.~Vasserman, B.~Hutchinson,
  E.~Spitzer, I.~D. Raji, and T.~Gebru.
\newblock Model cards for model reporting.
\newblock In \emph{Proceedings of the conference on fairness, accountability,
  and transparency}, pages 220--229, 2019.

\bibitem[Mnih and Hinton(2009)]{mnih2009scalable}
A.~Mnih and G.~E. Hinton.
\newblock {A Scalable Hierarchical Distributed Language Model}.
\newblock In \emph{Advances in Neural information Processing Systems}, pages
  1081--1088, 2009.

\bibitem[Mnih and Teh(2012)]{mnih2012fast}
A.~Mnih and Y.~W. Teh.
\newblock {A Fast and Simple Algorithm for Training Neural Probabilistic
  Language Models}.
\newblock \emph{Arxiv Preprint Arxiv:1206.6426}, 2012.

\bibitem[Morcos et~al.(2018)Morcos, Raghu, and Bengio]{morcos2018insights}
A.~S. Morcos, M.~Raghu, and S.~Bengio.
\newblock {Insights on Representational Similarity in Neural Networks with
  Canonical Correlation}, 2018.

\bibitem[Oord et~al.(2018)Oord, Li, and Vinyals]{oord2018representation}
A.~V.~D. Oord, Y.~Li, and O.~Vinyals.
\newblock {Representation Learning with Contrastive Predictive Coding}.
\newblock \emph{Arxiv Preprint Arxiv:1807.03748}, 2018.

\bibitem[Pearson(1901)]{pearson1901liii}
K.~Pearson.
\newblock {LIII. On Lines and Planes of Closest Fit to Systems of Points in
  Space}.
\newblock \emph{The London, Edinburgh, and Dublin Philosophical Magazine and
  Journal of Science}, 2\penalty0 (11):\penalty0 559--572, 1901.

\bibitem[Radford et~al.(2018)Radford, Narasimhan, Salimans, and
  Sutskever]{radford2018improving}
A.~Radford, K.~Narasimhan, T.~Salimans, and I.~Sutskever.
\newblock {Improving Language Understanding by Generative Pre-training}.
\newblock 2018.

\bibitem[Radford et~al.(2019)Radford, Wu, Child, Luan, Amodei, and
  Sutskever]{radford2019language}
A.~Radford, J.~Wu, R.~Child, D.~Luan, D.~Amodei, and I.~Sutskever.
\newblock {Language Models are Unsupervised Multitask Learners}.
\newblock \emph{Openai Blog}, 1\penalty0 (8), 2019.

\bibitem[Raghu et~al.(2017)Raghu, Gilmer, Yosinski, and
  Sohl-Dickstein]{raghu2017svcca}
M.~Raghu, J.~Gilmer, J.~Yosinski, and J.~Sohl-Dickstein.
\newblock {SVCCA: Singular Vector Canonical Correlation Analysis for Deep
  Learning Dynamics and interpretability}.
\newblock In \emph{Advances in Neural information Processing Systems}, pages
  6076--6085, 2017.

\bibitem[Roeder and Kingma(2019)]{roeder2019identifiability}
G.~Roeder and D.~P. Kingma.
\newblock {On the Identifiability of Representations in Supervised and
  Self-Supervised Learning}.
\newblock \emph{Conference On The Mathematical Theory of Deep Neural Networks},
  2019.

\bibitem[Sharif~Razavian et~al.(2014)Sharif~Razavian, Azizpour, Sullivan, and
  Carlsson]{sharif2014cnn}
A.~Sharif~Razavian, H.~Azizpour, J.~Sullivan, and S.~Carlsson.
\newblock {CNN Features Off-the-Shelf: An Astounding Baseline for Recognition}.
\newblock In \emph{Proceedings of The Ieee Conference On Computer Vision and
  Pattern Recognition Workshops}, pages 806--813, 2014.

\bibitem[Sohn(2016)]{sohn2016improved}
K.~Sohn.
\newblock {Improved Deep Metric Learning with Multi-class N-Pair Loss
  Objective}.
\newblock In \emph{Advances in Neural information Processing Systems}, pages
  1857--1865, 2016.

\bibitem[Sorrenson et~al.(2020)Sorrenson, Rother, and
  K{\"o}the]{sorrenson2020disentanglement}
P.~Sorrenson, C.~Rother, and U.~K{\"o}the.
\newblock Disentanglement by {{Nonlinear ICA}} with {{General
  Incompressible}}-flow {{Networks}} ({{Gin}}).
\newblock \emph{Arxiv:2001.04872 [Cs, Stat]}, Jan. 2020.

\bibitem[Vaswani et~al.(2017)Vaswani, Shazeer, Parmar, Uszkoreit, Jones, Gomez,
  Kaiser, and Polosukhin]{vaswani2017attention}
A.~Vaswani, N.~Shazeer, N.~Parmar, J.~Uszkoreit, L.~Jones, A.~N. Gomez,
  {\L}.~Kaiser, and I.~Polosukhin.
\newblock {Attention is All You Need}.
\newblock In \emph{Advances in Neural information Processing Systems}, pages
  5998--6008, 2017.

\bibitem[Vincent et~al.(2008)Vincent, Larochelle, Bengio, and
  Manzagol]{vincent2008extracting}
P.~Vincent, H.~Larochelle, Y.~Bengio, and P.-A. Manzagol.
\newblock {Extracting and Composing Robust Features with Denoising
  Autoencoders}.
\newblock In \emph{Proceedings of The 25th international Conference On Machine
  Learning}, pages 1096--1103, 2008.

\bibitem[Wolf et~al.(2019)Wolf, Debut, Sanh, Chaumond, Delangue, Moi, Cistac,
  Rault, Louf, Funtowicz, and Brew]{Wolf2019HuggingFacesTS}
T.~Wolf, L.~Debut, V.~Sanh, J.~Chaumond, C.~Delangue, A.~Moi, P.~Cistac,
  T.~Rault, R.~Louf, M.~Funtowicz, and J.~Brew.
\newblock {Huggingface's Transformers: State-of-the-art Natural Language
  Processing}.
\newblock \emph{Arxiv}, Abs/1910.03771, 2019.

\bibitem[Yang et~al.(2019)Yang, Dai, Yang, Carbonell, Salakhutdinov, and
  Le]{yang2019xlnet}
Z.~Yang, Z.~Dai, Y.~Yang, J.~Carbonell, R.~Salakhutdinov, and Q.~V. Le.
\newblock {XLNET: Generalized Autoregressive Pretraining for Language
  Understanding}.
\newblock \emph{Arxiv Preprint Arxiv:1906.08237}, 2019.

\end{thebibliography}
\bibliographystyle{abbrvnat}

\clearpage
\appendix
\section{Reproducing Experiments and Figures}
In this section, we present training and optimization details needed to reproduce our empirical validation of Theorem \ref{th:1}.
We also published notebooks and check-pointed weights for two crucial experiments that investigate the result in the small and massive scale regimes, for Figure \ref{fig:fig1} and GPT-2 (\url{https://github.com/google-research/google-research/tree/master/linear_identifiability}).
\label{app:train}
\subsection{Figure \ref{fig:fig1}}
\label{app:fig1}
We provide a Jupyter notebook and model checkpoints for reproducing Figure \ref{fig:fig1}.
Please refer to this for hyperparameter settings.
In short, we implemented a model \citep{mnih2012fast} in the family of Section \ref{sec:gdl} and trained it on the Billion Word dataset \citep{chelba2013one}.
This is illustrative of the property of Theorem \ref{th:1} because the relatively modest size of the parameter space (see notebook) and massive dataset minimizes model convergence and data availability restrictions, e.g., approaches the asymptotic regime.

The word embedding space is 2-D for ease of visualization.
We selected a subset of words, mapped them into their learned embeddings, and visualized them as points in the left and middle panes.
We then regress pane one onto pane two in order to learn the best linear transformation between them.
Note that if the two are linear transformations of each other, regression will recover that transformation exactly.

\subsection{Simulation Study: Classification by Deep Neural Networks}
Remaining training details are as follows.

Because our theory requires that the data generating process be expressible by the true generative model, we simulate this by training a 4 hidden layer MLP with two $2^6$ unit layers, and a 2-dimensional ``bottle neck" layer.
We optimize weights using Adam with a learning rate of $10^{-4}$ for $5*10^4$ iterations.

To make the classification problem more challenging, we additionally add 20 input dimensions of random noise.
The Adam optimizer with a learning rate of $3 \cdot 10^{-4}$ is used.

\subsection{Self-Supervised Learning for Image Classification}

To compute linear similarity between representations, we train two independent models in parallel.
For each model we define both $\rvfT$ and $\rvgT$ as a 3-layer fully connected neural network with $2^8$ units per layer and a fixed output dimensionality of $2^6$.
We define our model following Eq. \ref{eq:model}, where $S$ is the set of the other image patches from the current minibatch and optimize the objective of \citep{henaff2019data}.
We augment both sampled patches independently with randomized brightness, saturation, hue, and contrast adjustments, following the recipe of \citep{henaff2019data}.
We train on the CIFAR10 dataset \citep{krizhevsky2009learning} with batchsize $2^8$, using the Adam optimizer with a learning rate of $10^{-4}$ and the JAX~\citep{jax2018github} software package.
For each model, we early stop based on a validation loss failing to improve further.

Additional details about the experiments that generated Figure \ref{fig:synth}:
  \paragraph{Figure \ref{fig:synth} a.} Patches are sampled randomly from training images.
  \paragraph{Figure \ref{fig:synth} b.} For each model, we train for at most $3*10^4$ iterations, early stopping when necessary based on validation loss.
  \paragraph{Figure \ref{fig:synth} c.} For each model, we train for at most $3*10^4$ iterations, early stopping when necessary based on validation loss. 
  \paragraph{Figure \ref{fig:synth} d.} Error bars show standard error computed over 5 pairs of models after $1.5*10^4$ training iterations.

\subsection{GPT-2}
We include all details through a notebook in the code release. Pretrained GPT-2 weights as specified in the main text are publicly available from HuggingFace~\citet{Wolf2019HuggingFacesTS}.

\section{Proof that Linear Similarity is an Equivalence Relation}
\label{app:def1}
We claim that $\linsim$ is an equivalence relation.
It suffices to show that it is reflexive, transitive, and symmetric. 
\proof 
Consider some function $\rvgT$ and some $\bT^\prime, \bT^\star, \bT^\dagger \subset \Theta$.
Suppose $\bT^\prime \linsim \bT^\star$. 
Then, there exists an invertible matrix $\bb{B}$ such that $\rvg_{\bT^\prime}(\rvx) = \bb{B}\rvg_{\bT^\star}(\rvx)$.
Since $\rvg_{\bT^\star}(\rvx) = \bb{B}^{-1}\rvg_{\bT^\prime}(\rvx)$, $\linsim$ is symmetric. 
Reflexivity follows from setting $\rvg_{\bT^\star}$ to $\rvg_{\bT^\prime}$ and $\bb{B}$ to the identity matrix.
To show transitivity, suppose also that $\bT^\star \linsim \bT^\dagger$. Then, there exists an invertible $B$ such that $\rvg_{\bT^\star}(\rvx) = \bb{C} \rvg_{\bT^\dagger}(\rvx)$. Since $\rvg_{\bT^\prime} \linsim \rvg_{\bT^\star}$, $\bb{B}^{-1}\rvg_{\bT^\prime}(\rvx) = \bb{C} \rvg_{\bT^\dagger}(\rvx)$. Rearranging terms, $\rvg_{\bT^\prime}(\rvx) = \bb{B} \bb{C} \rvg_{\bT^\dagger}(\rvx)$, so that $\bT^\prime \linsim \bT^\dagger$ as required. 
\\ \qed

\section{Section \ref{sec:idanalysis} Continued: Case of Context Representation Function $\rvg$}
\label{sec:identifiability_g}

Our derivation of identifiability of $\rvgT$ is similar to the derivation of $\rvfT$.
The primary difference is that the normalizing constants in Equation (\ref{eq:normalizing}) do not cancel out.
First, note that we can rewrite \Eqref{eq:model} as:
\begin{align}
    \pT(\rvy | \rvx, \bS) = \exp(\widetilde{\rvfT}(\rvx, \bS)^\top \widetilde{\rvg}_{\bT}(\rvy))
\label{eq:model2}
\end{align}
where:
\begin{align}
    \widetilde{\rvfT}(\rvx, \bS) &= [-Z(\rvx, \bS); \rvfT(\rvx) ]\\
    \widetilde{\rvg}_{\bT}(\rvy) &= [1; \rvgT(\rvy)]\\
    Z(\rvx, \bS) &= \log \sum_{\rvy' \in \bS} \exp (\rvfT(\rvx)^\top \rvgT(\rvy')).
\end{align}

Below, we will show that for the model family defined in Section \ref{sec:gdl},
\begin{align}
    {p_{\bT^\prime}} = p_{\bT^*} \;\; \implies \;\; \rvg_{\bT^\prime}(\rvy) &= \bb{B}\,  \rvg_{\bT^\star}(\rvy),
\end{align}
where $\bb{B}$ is an invertible $(M \times M)$-dimensional matrix, concluding the proof of the linear identifiability of models in the family defined by Equation \ref{eq:model}.
We adopt the same shorthands as in the main text.

\subsection{Diversity condition}\label{sec:f_div}

We assume that for any $(\bT',\bT^*) \subset \Theta$ for which it holds that $p' = p^*$, and for any given $\rvy$, there exist $M+1$ tuples $\{(\rvx^{(i)}, \bS^{(i)})\}_{i=0}^{M}$, such that $\pD(\rvx^{(i)}, \rvy, \bS^{(i)}) > 0$, and such that the $((M+1) \times (M+1))$ matrices $\bb{M}'$ and $\bb{M}^*$ are invertible, where $\bb{M}'$ consists of columns $\widetilde{\rvf}^\prime(\rvx^{(i)}, \bS^{(i)})$, and $\bb{M}^*$ consists of columns $\widetilde{\rvf}^*(\rvx^{(i)}, \bS^{(i)})$.

This is similar to the diversity condition of Section \ref{sec:idanalysis} but milder, since a typical dataset will have multiple $\rvx$ for each $\rvy$.

\subsection{Proof}

With the data distribution $\pD(\rvx, \rvy, \bS)$, for a given $\rvy$, there exists a conditional distribution $\pD(\rvx, \bS | \rvy)$. Let $(\rvx, \bS)$ be a sample from this distribution. From \eqref{eq:model} and the statement to prove, it follows that:
\begin{align}
    p'(\rvy|\rvx, \bS) = p^*(\rvy|\rvx, \bS)
\end{align}
Substituting in the definition of our model from equation (\ref{eq:model2}), we find:
    \begin{align}
    \exp( \widetilde{\rvf}^\prime(\rvx, \bS)^\top  \widetilde{\rvg}^\prime(\rvy))
        =
        \exp( \widetilde{\rvf}^*(\rvx, \bS)^\top  \widetilde{\rvg}^*(\rvy)),
\end{align}
which, evaluating logarithms, becomes
\begin{align}
    \widetilde{\rvf}^\prime(\rvx, \bS)^\top  \widetilde{\rvg}^\prime(\rvy)
    = 
    \widetilde{\rvf}^*(\rvx, \bS)^\top  \widetilde{\rvg}^*(\rvy),
\label{eq:g}
\end{align}
which is true for any triple $(\rvx, \rvy, \bS)$ where $\pD(\rvy | \rvx, \bS) > 0$.

From $\bb{M}^\prime$ and $\bb{M}^*$ (Section \ref{sec:f_div}) and \eqref{eq:g} we form a linear system of equations, collecting the $M+1$ relationships together:
\begin{align}
    \bb{M}^{\prime^\top} \widetilde{\rvg}^\prime(\rvy) 
    &= 
    \bb{M}^{*\top}  \widetilde{\rvg}^*(\rvy) \label{eq:gtilde} \\
     \widetilde{\rvg}^\prime(\rvy)  &= \bb{A} \widetilde{\rvg}^*(\rvy),
\end{align}
where $\bb{A}=(\bb{M}^* \bb{M}^{\prime -1})^\top$, an invertible $(M+1) \times (M+1)$ matrix. 

It remains to show the existence of an invertible $M \times M$ matrix $\bb{B}$ such that
\begin{align}
    \rvg^\prime(\rvy) &= \bb{B} \rvg^*(\rvy).
\end{align}

We proceed by constructing $\bb{B}$ from $\bb{A}$. Since $\bb{A}$ is invertible, there exist $j$ elementary matrices  $\{ \bb{E}_1, \dots, \bb{E}_j\}$ such that their action ${\bb{R} = \bb{E}_j \bb{E}_{j-1} \dots \bb{E}_1}$ converts $\bb{A}$ to a (non-unique) row echelon form.
Without loss of generality, we build $\bb{R}$ such that the $a_{1,1}$ entry of $\bb{A}$ is the first pivot, leading to the particular row echelon form:
\begin{align}
\bb{R} \bb{A} = \left[ 
\begin{matrix} 
a_{1,1} & a_{1,2} & a_{1,3} & \dots & a_{1,m \times 1} \\
0 & \tilde{a}_{2,2} &  \tilde{a}_{2,3}  &\dots &  \tilde{a}_{2,m \times 1} \\
0 & 0 &  \tilde{a}_{3,3}  &\dots &  \tilde{a}_{2,m \times 1} \\
\vdots & \vdots &  \vdots & \ddots & \vdots \\
0 & 0 & \dots &  0 & \tilde{a}_{m\times 1,m \times 1} \\
\end{matrix}
\label{eq:Amatrix}
\right],
\end{align}
where $\tilde{a}_{i,j}$ indicates that the corresponding entry in $\bb{RA}$ may differ from $\bb{A}$ due to the action of $\bb{R}$.
Applying $\bb{R}$ to Equation (\ref{eq:gtilde}), we have 
\begin{align}
    \bb{R}\widetilde{\rvg}^\prime(\rvy)  &= \bb{RA} \widetilde{\rvg}^*(\rvy).
    \label{eq:RAmatrix}
\end{align}
We now show that removing the first row and column of $\bb {RA}$ and $\bb{R}$ generates matrices of rank $M$. Let $\overline{\bb {RA}}$ and $\overline{\bb{R}}$ denote the $(M \times M)$ submatrices formed by removing the first row and column of $\bb{RA}$ and $\bb{R}$ respectively.

Equation (\ref{eq:Amatrix}) shows that $\overline{\bb{RA}}$ has a pivot in each column, and thus has rank $M$. To show that $\overline{\bb{R}}$ is invertible, we must show that removing the first row and column reduces the rank of ${\bb{R} = \bb{E}_j \bb{E}_{j-1} \dots \bb{E}_1}$ by exactly 1. Clearly, each $\bb{E}_{k}$ is invertible, and their composition is invertible. We must show the same for the composition of $\overline{\bb{E}_{k}}$.  

There are three cases to consider, corresponding to the three unique types of elementary matrices. Each elementary matrix acts on $\bb{A}$ by either (1) swapping rows $i$ and $j$, (2) replacing row $j$ by a multiple $m$ of itself, or (3) adding a multiple $m$ of row $i$ to row $j$. We denote elementary matrix types by superscripts.

In Case (1), $\bb{E}_{k}^{1}$ is an identity matrix with row $i$ and row $j$ swapped. For Case (2), $\bb{E}_{l}^{2}$ is an identity matrix with the $j,j^{th}$ entry replaced by some $m$. For each $\bb{E}_{k}^{1}$ and $\bb{E}_{l}^{2}$ in $\bb{R}$ , where $1\leq k,l \leq j$, we know that the indices $i, j \geq 2$, because we chose the first entry of the first row of $\bb{A}$ to be the pivot, and hence do not swap the first row, or replace the first row by itself multiplied by a constant. This implies that removing the first row and column of $\bb{E}_{k}^{1}$ and $\bb{E}_{l}^{2}$ removes a pivot entry 1 in the $(1,1)$ position, and removes zeros elsewhere. Hence, the $(M \times M)$ submatrices $\overline{\bb{E}_{k}^{1}}$ and $\overline{\bb{E}_{l}^{2}}$ are elementary matrices with rank $M$. 

For Case (3), $\bb{E}_{k}^{3}$ has some value $m \in \mathbb{R}$ in the $j,i^{th}$ entry, and 1s along the diagonal. In this case, we may find a non-zero entry in some $\bb{E}_{k}^{3}$, so that, e.g., the second row has a pivot at position $(2,2)$. Without loss of generality, suppose $i=1$, $j = 2$ and let $m$ be some nonzero constant. Removing the first row and column of $\bb{E}_{1}^{3}$ removes this $m$ also. Nevertheless, $\overline{\bb{E}_{1}^3} = \bb{I}_M$, the rank $M$ identity matrix. For any other $\bb{E}_{k}^{3}$ $1 < i \leq M+1$, $j \geq 2$ because we chose $a_{1,1}$ as the first pivot, and hence do not swap the first row, or replace the first row by itself multiplied by a constant. In both cases, removing the first row and first column creates an $\overline{\bb{E}_{k}^{3}}$ that is a rank $M$ elementary matrix.

We have shown by the above that $\overline{\bb{R}}$ is a composition of rank $M$ matrices. 
We now construct the matrix $\bb{B}$ by removing the first entries of $\widetilde{\rvg}^\prime$ and $\widetilde{\rvg}^\star$, and removing the first row and first column of $\bb{R}$ and $\bb{RA}$ in Equation (\eqref{eq:RAmatrix}). Then, we have
\begin{align}
     \overline{\bb{R}} \rvg^\prime(\rvy)  &= \overline{\bb{RA}}\rvg^*(\rvy), \\
     \rvg^\prime(\rvy)  &= \overline{\bb{R}}^{-1}\overline{\bb{RA}}\rvg^*(\rvy).
\end{align}
Choosing $\bb{B} = \overline{\bb{R}}^{-1}\overline{\bb{RA}}$ proves the result.\\
\qed

\section{Reductions to Canonical Form of Equation (\ref{eq:model})}
\label{app:gdl_equations}
In the following, we show membership in the model family of \Eqref{eq:model} using the mathematical notation of the papers under discussion in Section \ref{sec:lit}. Note that each subsection will change notation to match the papers under discussion, which varies quite widely. We employ the following colour-coding scheme to aid in clarity:
\begin{align*}
    \log \pT(\rvy | \rvx, \bS) &= \dcolbox{aaaa}{\rvfT(\rvx)}^\top \ccolbox{bbbb}{\rvgT(\rvy))}
     - \largeconstcolbox{cccc}{\sum_{\rvy' \in \bS} \exp (\rvfT(\rvx)^\top \rvgT(\rvy'))}, \\
\end{align*}
where $\rvfT(\rvx)$ is generalized to a $\dcolbox{aa}{\text{data representation function}}$, $\rvgT(\rvy)$ is generalized to a $\ccolbox{bb}{\text{context representation function}}$, and $\sum_{\rvy' \in \bS} \exp (\rvfT(\rvx)^\top \rvgT(\rvy'))$ is some $\constcolbox{cc}{\text{constant}}$.

\subsection{CPC}
Formally, consider a sequence of points $\rvx_t$.  
We wish to learn the parameters $\cpaparams$ to maximize the $k$-step ahead predictive distribution $p(\rvx_{t+k}|\rvx_t, \cpaparams)$.
In the image patch example, each patch center $i,j$ is indexed by $t$.
Each $\rvx_t$ is mapped to a sequence of feature vectors $\vz_t = f_{\theta}(\rvx_{t})$
An autoregressive model, already updated with the previous latent representations $\vz_{\leq t-1}$, transforms the $\vz_t$ into a ``context" latent representation $\rvc_t = \ga(\vz_{\leq t})$.
Instead of predicting future observations $k$ steps ahead, $\xtk$, directly through a generative model $p_k(\xtk | \rvc_t)$, \citet{oord2018representation} model a density ratio in order to preserve the mutual information between $\xtk$ and $\rvc_t$.

\noindent \paragraph{Objective} 
Let $\bb{X}=\{\rvx_1, \dots, \rvx_N\}$ be a set of $N$ random samples containing one positive sample from $p(\xtk |\rvc_t)$ and $N-1$ samples from the proposal distribution $p(\xtk)$.
 \citet{oord2018representation} define the following link function: $l_k(\xtk, \rvc_t)  \triangleq \exp \left(\rvz_{t+k}^\top \bb{W}_k \rvc_t \right)$. Then, CPC optimizes
 
\begin{align} 
 -\E_{\bb{X}} \left[ \log \frac{l_k(\xtk, \rvc_t)}{\sum_{x_j \in X} l_k( \rvx_j, \rvc_t)} \right] 
    = -\E_{\bb{X}} \left[ \log \frac{\exp \left( 
    \tikzmarkin[set fill color=\ccol, set border color=\ccol]{a}(.05,-0.125)(-.05,0.3)
    \rvz_{t+k}
    \tikzmarkend{a}
    ^\top 
    \tikzmarkin[set fill color=\dcol, set border color=\dcol]{b}(.05,-0.125)(-.05,0.3)
    \bb{W}_k \rvc_t
    \tikzmarkend{b}
    \right)}{
    \tikzmarkin[set fill color=\constcol, set border color=\constcol]{c}(.05,-0.2)(-.05,0.30)
    \sum_{\rvx_j \in \bb{X}} \exp \left( \rvz_{j}^\top \bb{W}_k \rvc_t\right) 
    \tikzmarkend{c}
    }  \right].
    \label{eq:cpc_loss}
\end{align} 
Substituting in the definition of $l_k$ makes equation (\ref{eq:cpc_loss}) identical to the model family (\Eqref{eq:model}).

\subsection{Autoregressive language models (e.g. GPT-2)}
Let $\gU = \{u_1, \dots, u_n\}$ be a corpus of tokens.
Autoregressive language models maximize a log-likelihood $L(\gU) = \sum_{i=1}^n \log P(u_i | u_{i-k}, \dots, u_{i-1}; \Theta)$,
Concretely, the conditional density is modelled as
\begin{align*}
    &\log P(u_i | u_{i-k:i-1}; \Theta) \\  &\qquad \qquad = \ccolbox{h}{\bb{W}_{i:}} \dcolbox{g}{\rvh_i} - \largeconstcolbox{i}{\log \sum_{j} \exp( \bb{W}_{j:}\rvh_i  )},
\end{align*} 
where $\rvh_i$ is the $m\times 1$ output of a function approximator (e.g. a Transformer decoder \citep{liu2018generating}), and $\bb{W}_{i:}$ is the $i$'th row of the $|\gU| \times m$ token embedding matrix. 

\subsection{BERT}
Consider a sequence of text $\rvx = [x_1, \dots, x_T]$. 
Some proportion of the symbols in $\rvx$ are extracted into a vector $\bar{\rvx}$, and then set in $\rvx$ to a special null symbol, ``corrupting" the original sequence.
This operation generates the corrupted sequence $\ubar{\rvx}$.
The representational learning task is to predict $\bar{\rvx}$ conditioned on $\ubar{\rvx}$, that is, to maximize w.r.t. $\bT$:
\begin{align*}
  \log p_{\theta}(\bar{\rvx} | \ubar{\rvx}) 
  \approx \sum_{t=1}^T m_t \log p_{\theta}(x_t | \ubar{\rvx}) = \sum_{t=1}^T m_t \Bigg( 
  \dcolbox{j}{H_{\theta}(\ubar{\rvx})_t}^\top \ccolbox{k}{e(x_t)} 
- \largeconstcolbox{l}{\log \sum_{x'} \exp \left(H_{\theta}(\ubar{\rvx})_t^\top e(x') \right)} ~\Bigg),
\end{align*} 

where $H$ is a transformer, $e$ is a lookup table, and $m_t = 1$ if symbol $x_t$ is masked. That is, corrupted symbols are ``reconstructed" by the model, meaning that their index is predicted. 
As noted in \citet{yang2019xlnet}, BERT models the joint conditional probability $p(\bar{\rvx}  |\ubar{\rvx})$ as factorized so that each masked token is separately reconstructed. 
This means that the log likelihood is approximate instead of exact.

\subsection{QuickThought Vectors}
Let $\rvf$ and $\rvg$ be  functions that take a sentence as input and encode it into an fixed length vector.
Let $s$ be a given sentence, and $S_{ctxt}$ be the set of sentences appearing in the context of $s$ for a fixed context size. Let $S_{cand}$ be the set of candidate sentences considered for a given context sentence $s_{ctxt} \in S_{ctxt}$. Then, $S_{cand}$ contains a valid context sentence $s_{ctxt}$ as well as many other non-context sentences. $S_{cand}$ is used for the classification objective.
For any given sentence position in the context of $s$ (for example, the preceding sentence), the probability that a candidate sentence $s_{cand}\in S_{cand}$ is the correct sentence for that position is given by
\begin{align*}
    \log p(s_{cand} | s, S_{cand}) =  \ccolbox{q}{f_{\theta}(s)}^\top \dcolbox{r}{g_{\theta}(s_{cand}))} - \largeconstcolbox{s}{\log \sum_{s' \in S_{cand}}\exp \left(f_{\theta}(s)^\top g_{\theta}(s'_{cand})\right)}.
\end{align*}

\subsection{Deep Metric Learning}
\label{sec:deep_metric}
The {\it multi-class N-pair loss} in \citet{sohn2016improved}  is proportional to
\begin{align*}
    \log N -\frac{1}{N}\sum_{i=1}^N \log \left( 1 + \sum_{j \neq i} \exp\{ \rvfT(x_i)^\top \rvfT(y_j) - \rvfT(x_i)^\top \rvfT(y_i))\} \right), 
\end{align*} 
    which can be simplified as
    \begin{align*}
   & -\frac{1}{N}\sum_{i=1}^N \log \left( \frac{1}{K}\sum_{j =1}^K \exp\{\rvfT(x_i)^\top \rvfT(y_j) - \rvfT(x_i)^\top \rvfT(y_i)\} \right) \\
    &= \frac{1}{N}\sum_{i=1}^N \log \left( \frac{1}{\frac{1}{K}\sum_{j =1}^K \exp\{\rvfT(x_i)^\top \rvfT(y_j) - \rvfT(x_i)^\top \rvfT(y_i)\}} \right) \\
    &= \frac{1}{N}\sum_{i=1}^N \log \left(  \frac{\exp \{ \rvfT(x_i)^\top \rvfT(y_i) \}}{\frac{1}{K}\sum_{j =1}^K \exp\{\rvfT(x_i)^\top \rvfT(y_j)\}} \right).
\end{align*}
Setting N to 1 and evaluating the log gives
\begin{align*}
    \dcolbox{qa}{\rvfT(x_i)}^\top \ccolbox{qb}{\rvfT(y_i)}  - \largeconstcolbox{qd}{\frac{1}{K}\sum_{j =1}^K \exp(\rvfT(x_i)^\top \rvfT(y_j))},
\end{align*}
which is \Eqref{eq:model} where $\rvfT=\rvgT$.
\subsection{Neural Probabilistic Language Models (NPLMs)}
\label{sec:nplm}
Figure \ref{fig:fig1} is a neural probabilistic language model as proposed in \citet{mnih2012fast}.
\citet{mnih2012fast} propose using a log-bilinear model \citep{mnih2009scalable} which, given some context $h$, learns a context word vectors $r_w$ and target word vectors $q_w$.
Two different embedding matrices are maintained, in other words: one to capture the embedding of the word and the other the context.
The representation for the context vector, $\hat{q}$, is then computed as the linear combination of the context words and a context weight matrix $C_i$ so that $\hat{q} = \sum_{i=1}^{n-1}C_i r_{w_i}$.
The score for the match between the context and the next word is computed as a dot product, e.g., $\score = \hat{q}^\top \tilde{q}_w$\footnote{We have absorbed the per-token baseline offset $b$ into the $q_w$ defined in \citet{mnih2012fast}, forming the vector $\tilde{q}_w$ whose $i$'th entry is $(q_w)_i = (q_w)_i + b_w / (\hat{q})_i$} and substituting into the definition of $P_{\theta}^h(w)$, we see that
\begin{align*}
\log \nlm = 
    \dcolbox{d}{\hat{q}}^\top \ccolbox{e}{\tilde{q}_w}
    - \largeconstcolbox{f}{\log \sum_{w'} \exp \left( \hat{q}^\top \tilde{q}_{w'} \right)}
\end{align*}
shows that \citet{mnih2012fast} is a member of the model family.

Interestingly, a touchstone work in the area of NPLMs, Word2Vec \citep{mikolov2013distributed}, does not fall under the model family due to an additional nonlinearity applied to the score of \citet{mnih2012fast}.
\end{document}